\documentclass{SCIS2022}
\usepackage{amsmath,amsfonts}
\usepackage{algorithmic}
\usepackage{algorithm}
\usepackage{array}
\usepackage{textcomp}
\usepackage{stfloats}
\usepackage{url}
\usepackage[utf8]{inputenc}
\usepackage[T1]{fontenc}
\usepackage{CJKutf8}
\usepackage{booktabs}
\usepackage{threeparttable}
\usepackage{multirow}
\usepackage{verbatim}
\usepackage{graphicx}
\usepackage{cite}
\usepackage{caption}

\begin{document}
\ArticleType{RESEARCH PAPER}
\Year{2022}
\Month{}
\Vol{}
\No{}
\DOI{}
\ArtNo{}
\ReceiveDate{}
\ReviseDate{}
\AcceptDate{}
\OnlineDate{}

\title{A optimization framework for herbal
prescription planning based on deep
reinforcement learning}

\author[1+]{Kuo Yang}{}
\author[1+]{Zecong Yu}{}
\author[1]{Xin Su}{}
\author[1]{Xiong He}{}
\author[1]{Ning Wang}{}
\author[1]{Qiguang Zheng}{}
\author[1]{\\Feidie Yu}{}
\author[2]{Zhuang Liu}{}
\author[3]{Tiancai Wen}{}
\author[1]{Xuezhong Zhou}{{xzzhou@bjtu.edu.cn}}
\AuthorMark{Author A}

\AuthorCitation{Author A, Author B, Author C, et al}


\address[1]{School of Computer and Information Technology, Beijing Jiaotong University, Beijing, {\rm 100044}, China}
\address[2]{Guangzhou University of Chinese Medicine, Guangzhou, {\rm 510006}, China}
\address[3]{Data Center of Traditional Chinese Medicine, China Academy of Chinese Medical Sciences, Beijing, {\rm 100700}, China}

\abstract{Treatment planning for chronic diseases is a critical task in medical artificial intelligence, particularly in traditional Chinese medicine (TCM). However, generating optimized sequential treatment strategies for patients with chronic diseases in different clinical encounters remains a challenging issue that requires further exploration. In this study, we proposed a TCM herbal prescription planning framework based on deep reinforcement learning for chronic disease treatment (PrescDRL). PrescDRL is a sequential herbal prescription optimization model that focuses on long-term effectiveness rather than achieving maximum reward at every step, thereby ensuring better patient outcomes. We constructed a high-quality benchmark dataset for sequential diagnosis and treatment of diabetes and evaluated PrescDRL against this benchmark. Our results showed that PrescDRL achieved a higher curative effect, with the single-step reward improving by 117\% and 153\% compared to doctors. Furthermore, PrescDRL outperformed the benchmark in prescription prediction, with precision improving by 40.5\% and recall improving by 63\%. Overall, our study demonstrates the potential of using artificial intelligence to improve clinical intelligent diagnosis and treatment in TCM.}


\maketitle

\section*{Abbreviations}
TCM, traditional Chinese medicine; DDTS, dynamic diagnosis and treatment scheme; PrescDRL, herbal prescription planning based on deep reinforcement learning; RL, reinforcement learning; MDP, Markov decision process; STC, sequence termination condition; SSR, single-step return; SCR, single-step cumulative return; MCR, multi-step cumulative return; HPC, herbal prescription cluster; HPTP, herbal prescription treatment planning.

\section{Introduction}
Intelligent diagnosis and automatic drug recommendation have become important topics in medical artificial intelligence\cite{patel2009coming}. The optimization problem of dynamic diagnosis and treatment scheme (DDTS) considers a patient's treatment as a sequential decision-making process\cite{alagoz2010markov}, aiming to identify the best sequential treatment schema\cite{deng2021application}. In the field of Traditional Chinese Medicine (TCM)\cite{li2015mapping}, DDTS optimization typically requires consideration of a patient's status (e.g., symptoms and signs) at each stage, and generates an herbal prescription treatment plan (HPTP). TCM doctors obtain a patient's symptom descriptions and corresponding syndromes through the "Four Examinations" method of "watching, listening, asking, and feeling"\cite{cui2018diagnosis}. Unlike prescription recommendation, which predicts the appropriate prescription based on a patient's current situation, DDTS optimization focuses on providing the best treatment at each stage to maximize the treatment effect of the entire sequence and identify the best sequential decision-making path. DDTS optimization prioritizes the outcome of a sequential treatment process rather than the outcome of a particular treatment.

With the explosive development of deep learning technology, it has gradually been applied to a variety of biomedical problems, such as disease gene prediction\cite{yang2020pdgnet,yang2018heterogeneous}, drug target prediction\cite{zhang2023drugai}, drug repositioning\cite{yang2023dronet}. Since the appearance of AlphaGO\cite{silver2016mastering} in 2015, deep reinforcement learning (DRL) has emerged as a research hotspot in medical artificial intelligence, combining the depth perception of deep learning\cite{lei2021deep} with the decision-making of reinforcement learning (RL) to achieve optimal decision-making control\cite{li2018deep}. Many excellent diagnosis and treatment planning models based on RL have been proposed by researchers in recent years\cite{nemati2016optimal,padmanabhan2019optimal,ghassemi2018personalized,lin2018deep,raghu2017deep,raghu2017continuous,raghu2018model,futoma2018learning,lopez2019deep}. For example, Shamim et al. proposed a circular decision-making framework based on RL, which provides personalized dose schemes for patients\cite{nemati2016optimal}. Liu et al. constructed an RL model for the prevention and treatment of graft-versus-host disease in leukemia patients\cite{liu2017deep}. Wang et al. proposed a supervised RL model based on cyclic neural networks to recommend dynamic diagnosis and treatment schemes\cite{wang2018supervised}. In the field of Traditional Chinese Medicine (TCM), Feng proposed the use of a partially observable Markov decision process (POMDP) model to mine the optimal DDTS\cite{FengQ2011}. Hu proposed a deep RL algorithm framework for optimizing the sequential diagnosis and treatment scheme of TCM\cite{HuX2019}.

With the intricate mechanisms of herb combinations in prescriptions, combined diseases in patients, and individual differences among patients, designing an appropriate DDTS optimization model remains a challenge\cite{gijsen2001causes}. Current RL-based DDTS optimization algorithms, on one hand, do not effectively learn from the medication rules of experienced TCM doctors and fail to achieve satisfactory results. On the other hand, they do not fully represent the patient's state space and action space. Consequently, there is a pressing need for more accurate and dependable models to enhance the practicality of auxiliary diagnosis and treatment and to recommend more reliable HPTP for patients.

With the availability of large-scale real-world clinical data\cite{zhang2012real} and advancements in artificial intelligence\cite{russell2002artificial}, it is now possible to construct robust computational models for recommending appropriate prescriptions\cite{coudray2018classification}. Three main categories of prescription recommendation methods have emerged, including traditional machine learning-based\cite{wang2019tcm,wu2022bayesian}, topic model-based\cite{yao2018topic,zhang2011topic}, and deep learning-based methods\cite{jin2021kg,li2021kgapg,li2019exploration}.For instance, Li et al.\cite{li2019exploration} proposed an improved seq2seq model to generate herbal prescriptions, while Yu et al.\cite{hu2019automatic} developed a model based on CNN and topic model to predict TCM prescriptions. Liao et al.\cite{liao2019convolutional} proposed a CNN-based model that extracts facial image features and maps the relationship between facial features and drugs to predict herbal prescriptions. Zhou et al.\cite{zhou2021fordnet} proposed an effective formula recommendation framework called FordNet, which integrates macro and micro information using deep neural networks. Dong et al.\cite{dong2021tcmpr} proposed a subnetwork-based symptom term mapping method (SSTM), and constructed a SSTM-based TCM prescription recommendation method TCMPR. Despite the growing number of studies on herbal prescription recommendation, a significant challenge remains in bridging the gap between treatment planning based on reinforcement learning and recommending specific prescriptions

\begin{figure}[!h]
\centering 
\includegraphics[width=\linewidth]{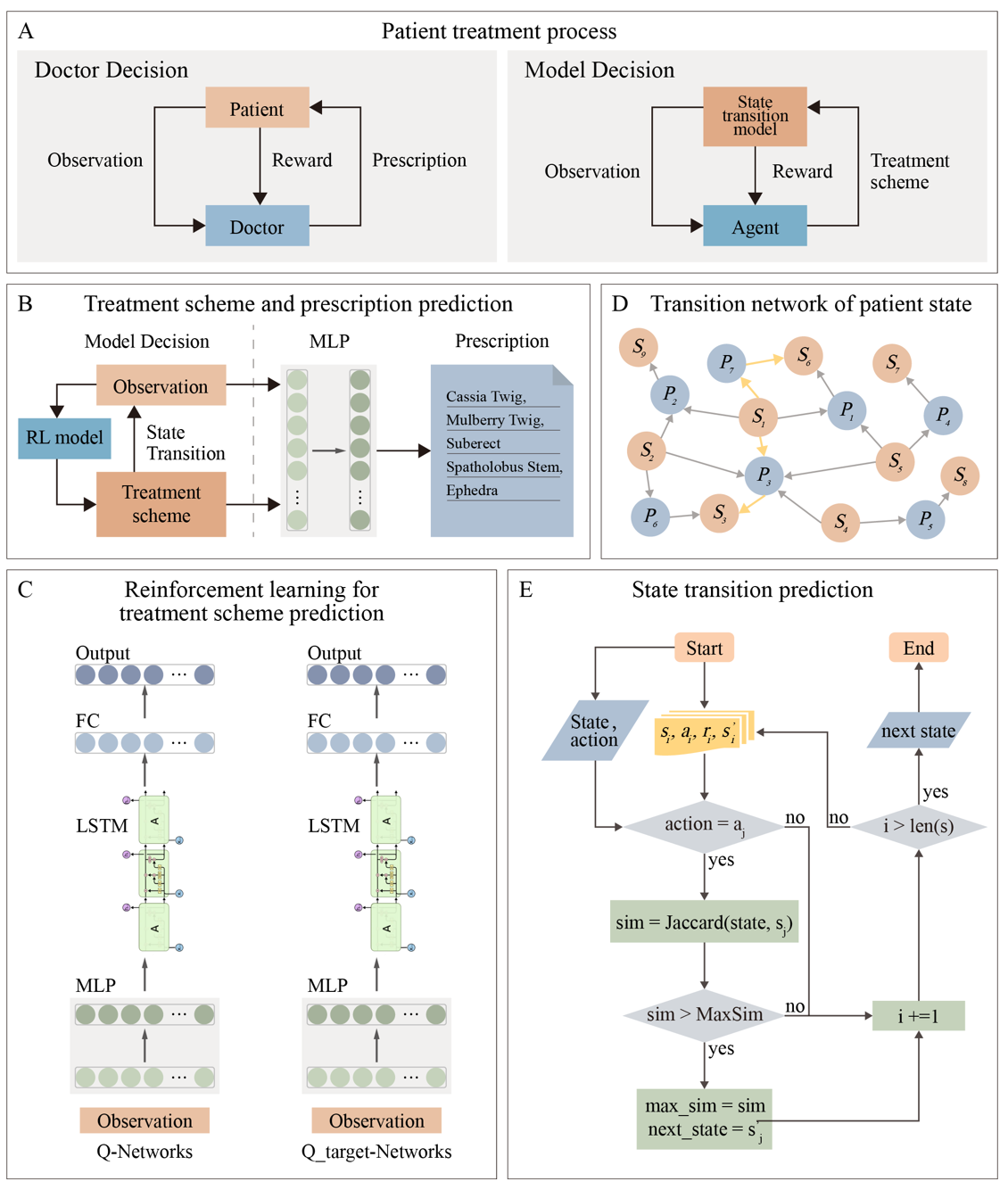}
\caption{Overall framework of PrescDRL. (A) Diagnose and treatment process of doctors and intelligence decision model. (B) The macro framework of PrescDRL. (C) RL-based prediction module of diagnose and treatment scheme. (D) Transition network of patient states. (E) The prediction module of state transition.}
\label{fig:1}
\end{figure}

In our study, we present PrescDRL, a novel model for optimizing diagnosis and treatment schemes using deep reinforcement learning (Figure~\ref{fig:1}A-\ref{fig:1}E). Initially, we constructed a high-quality benchmark dataset for sequential diagnosis and treatment of diabetes, and subsequently designed the PrescDRL framework for herbal prescription treatment planning. Unlike traditional reward-driven approaches, PrescDRL focuses on long-term effectiveness to ensure better outcomes for patients. We formulated the optimization of Diagnosis and Treatment Treatment Scheme (DDTS) as a reinforcement learning task, with patient symptom observations as inputs and High-Performance Treatment Plan (HPTP) as the optimization goal. We then employed a multi-layer neural network to predict TCM prescriptions using patient symptoms and recommended HPTP as inputs. Finally, the recommended HPTP and herbal prescriptions are proposed to patients as a treatment scheme. Moreover, PrescDRL includes a prediction module for TCM prescription based on patient symptoms and HPTP. Our comprehensive experiments demonstrate that PrescDRL outperforms doctors in providing HPTP with better expected effectiveness and has a higher prediction performance for TCM prescription.

\section{MATERIALS AND METHODS}
\subsection{Clinical sequential data of diabetes}
In this section, we present a benchmark dataset of clinical sequential diagnosis and treatment for diabetes, which serves as an example to train the optimization method of DDTS. (Ethics approval of this study has been obtained from ethics committee of institute of Clinical Basic Medicine of Traditional Chinese Medicine (NO. 2016NO.11-01)). In this dataset, the symptom observations of patients are selected as the states, and herbal prescription prescribed by doctors as actions in reinforcement learning (RL) model.

\begin{figure}[ht]
\centering
\includegraphics[width=\linewidth]{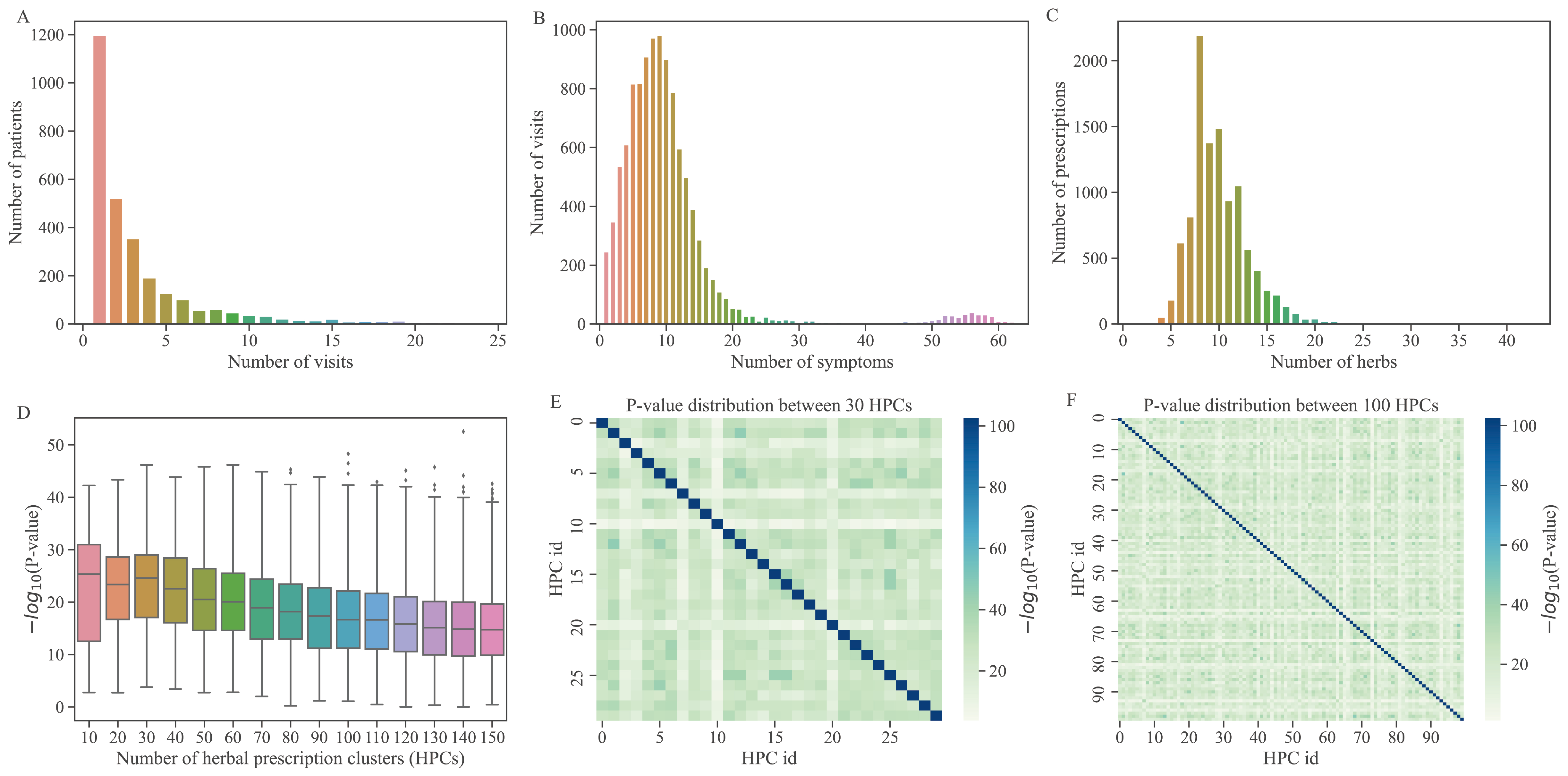}
\caption{Distribution of sequential diagnosis and treatment data. (A) Distribution of the number of patient visits. (B) Distribution of the number of patients’ symptoms. (C) Distribution of the number of the herbs in prescriptions. (D) P-values distribution of different number of herbal prescriptions clusters. (E) P-value distribution of 30 prescription clusters. (F) P-value distribution of 100 prescription clusters.}
\label{fig:2}
\end{figure}

To construct a standard dataset for sequential decision-making in TCM, we first extracted 10,666 medical records of 2,895 diabetic patients from Guang’anmen Hospital. As depicted in Figure~\ref{fig:2}A, 49.6\% of the patients had only one medical record and each patient had an average of 3.68 medical records. For each medical record, we extracted the patient's symptoms and an herbal prescription consisting of multiple herbs for treatment. With the exception of 334 medical records with over 40 symptoms, the number of symptoms per patient was normally distributed (Figure~\ref{fig:2}B), with an average of 10.386 symptoms per medical record. Similarly, the number of herbs per prescription was normally distributed (Figure~\ref{fig:2}C), with an average of 10.059 herbs per prescription. We screened 1459 patients with more than one medical visit and obtained 5,638 medical records, which were arranged into diagnosis and treatment sequences based on clinic time.

\subsection{A deep reinforcement learning framework to optimize herbal prescription treatment planning}

The optimization of DDTS is essentially a Markov Decision Process (MDP, \cite{bellman1957markovian}). To tackle this problem, we propose an optimization model for herbal prescription treatment planning based on two high-performance deep RL models, namely, DRN \cite{mnih2013playing} and DRQN \cite{hausknecht2015deep}. DQN is a combination of Q-learning and convolutional neural network that can perform RL tasks. On the other hand, DRQN first extracts features using two fully connected layers, followed by a LSTM layer, and then predicts the action value using a final fully connected layer (Figure \ref{fig:1}C).

In the RL framework, the agent acts as a virtual intelligent doctor, with the patient's state serving as the environment and prescribing herbal medication to the patient as the agent's action. The key components of RL models are defined as follows: 1) The state space is denoted by $S$, where a state $s \in S $ represents the observation of a patient's symptoms; 2) The action space is denoted by $A$, where an action $a\in A$ represents the herbal medication prescribed to the patient; 3) The reward function is denoted by $R(s,a)$, which returns a reward after the agent takes action in state $s$; 4) The virtual environment is denoted by $E$, which is an offline virtual environment based on sequential clinical data; 5) The state transition is denoted by $T$, where each transition is obtained using a prediction strategy.

\subsubsection{The state observations of patients}

In the DDTS optimization problem, the patient's state is a key component of the reinforcement learning model. In TCM clinics, doctors obtain symptom descriptions of patients through "seeing, hearing, asking, and cutting" \begin{CJK}{UTF8}{gbsn}(“望闻问切”)\end{CJK}, summarize the syndrome type, and prescribe appropriate treatments. However, since the true state of the patient is not available, even experienced doctors cannot fully determine the specific conditions of patients. Therefore, the patient's symptoms observed by the doctors are used to approximate the patient's state.

In the diabetes dataset, the distribution of patient symptoms (Figure~\ref{fig:2}B) shows that the number of symptoms varies among patients (average of 10 symptoms per patient). The core symptoms for each disease typically differ, and different symptoms may have varying importance. However, it is challenging to obtain a precise symptom grading for diabetes, and thus different symptoms are typically considered to have equal weight. As a result, a patient's state is represented by a symptom vector, where the symptoms present in the patient are marked as 1 and those that are not are marked as 0.

\subsubsection{The action spaces of virtual doctor}

In TCM diagnosis and treatment, doctors prescribe herbal prescriptions based on the patient's symptoms. From all the medical records, we obtained 9,695 distinct herbal prescriptions. Considering all these prescriptions as actions of the RL model would greatly increase the difficulty of training and convergence of RL algorithms due to the large number of actions. Therefore, it is necessary to reduce the number of actions by converting prescription numbers into a suitable discrete space. This will reduce the model complexity and improve the convergence speed.

To reduce the number of prescriptions, we employed the K-means clustering algorithm\cite{bellman1957markovian} to cluster these prescriptions and used prescription's herb information as the feature. We performed a parameter tuning experiment to obtain a proper number of herbal prescription clusters (HPC) which is considered a hyperparameter. We tested different values of HPC ranging from 30 to 150 with increments of every 10 categories. A good HPC result is expected to have different categories with significantly different herbs. To achieve this, we used the Chi-square test\cite{pearson1900x} to calculate the statistical difference between any two clusters based on the composition of herbs prescribed in different clusters. The resulting HPC is used as the action of the RL models.

\subsubsection{Design of reward function}

The aim of RL-based DDTS optimization is to use a vast number of medical records to predict the optimal sequence of herbal prescriptions for patients. The objective is not only to maximize the treatment effect but also to ensure that the predicted prescriptions are reasonable. This means that the efficacy of the predicted prescriptions should be within a reasonable range, and they should not have side effects on patients or contradict drug indications. 

Due to the absence of curative effect evaluation data in the diabetes dataset, we utilized the change in symptom scores between two consecutive visits before and after treatment as the immediate reward value for the current patient action. Additionally, we calculated the Jaccard coefficient to measure the similarity between the predicted action and the actual prescription provided by the doctor. A higher reward value was assigned to actions that had a higher similarity to the doctor's prescription. Therefore, the reward function was formulated as follows:

\begin{equation}
\mathcal{R}(s, a)=\gamma \sum_{i=1}^{n} \alpha_{i}\left(s_{i}-s_{i}^{\prime}\right)+\beta J a c\left(a, a^{\prime}\right)
\end{equation}

\begin{equation}
\operatorname{Jac}\left(a, a^{\prime}\right)=\frac{\left|a \cap a^{\prime}\right|}{\left|a \cup a^{\prime}\right|}
\end{equation}
where $\alpha_i$ represents the weight of patient's $i-th$ symptom, $s_i$ represents the $i-th$ symptom of the patient at the current visit, and $s_i^{\prime}$ represents the $i-th$ symptom of the patient at the next visit. $\gamma $ denotes the weight of therapeutic effect of patients, $\beta$ denotes the weight of risk, a denotes the prescription given by the doctor, and $a^{\prime}$ denotes the prescriptions predicted by the model.

\subsubsection{Virtual environment construction}
To overcome the impossibility of training the proposed DDTS optimization model in the real diagnosis and treatment process, we developed an off-line virtual environment based on the available medical records of patients. We constructed a tetrad, represented as $(s_1,a,r,s_2)$, using the symptom observation and prescriptions of each patient in the current and next diagnosis and treatment. In this tetrad, $s_1$ denotes the current symptom observation of the patient, denotes the action based on $s_1$, $r$ denotes the reward received after performing the action $a$, and $s_2$ denotes the new symptom observation of the patient after the action $a$. We obtained 4,179 tetrads from the medical records, which served as a virtual environment to train the deep RL model.


\subsubsection{State transition prediction and termination}

In the optimization of DDTS with the deep RL model, one of the main challenges is obtaining the next symptom observations after conducting an action based on the current symptoms due to the lack of tetrads constructed in the training stage. To address this issue, we utilized the state transition network, which includes states and actions (Figure \ref{fig:1}D), to predict the patient's symptoms after treatment. Specifically, we developed a prediction strategy that involves screening out all tetrads $(s_1,a,r,s_2 )$ with the same predicted action in the training set, calculating the Jaccard similarity between $s_1$ the symptom observations in each tetrad and the current symptoms, and selecting the $s_1$ tetrad with the highest similarity to the current symptoms. Finally, $s_2$ in the same tetrad as $s_1$ is selected as the state of the patient after treatment.

The distribution of symptoms in patients (Figure~\ref{fig:2}B) indicates that 94.7\% of patients have between 1 and 20 symptoms. Based on TCM expert recommendations and the symptom distribution, we define the first sequence termination condition (STC) as a patient's symptom score $\le $3. According to the evaluation criteria of diabetes treatment effect, a 30\% reduction in symptom score is considered effective, while a 70\% reduction is considered markedly effective. Therefore, the second STC is defined as a 60\% reduction in the patient's symptom score. The distribution of consultations (Figure~\ref{fig:2}A) shows that 93\% of patients have between 1 and 10 consultations (average number is 3.7). The last STC is number of iterations bigger than 15.

\subsection{A multi-layer neural network for herbal prescription recommendation}

In clinical practice of TCM, the ultimate goal of intelligent decision-making for diagnosis and treatment is to recommend effective herbal prescriptions to patients. By utilizing the trained deep RL models, we can obtain the sequential HPC for patients. In order to predict appropriate prescriptions, we model the prescription recommendation as a task of multi-label prediction. To achieve this, we constructed a multi-layer neural network (i.e., multi-layer perception), which takes the patient's symptoms and the HPC predicted by the RL models as input features, and outputs the predicted herbal prescription (Figure~\ref{fig:1}B).

\subsection{Experimental Design}

\subsubsection{Parameter setting}
In the DDTS optimization experiment, we used a total of 1,495 patient samples, of which 80\% (1,203 samples) were used for training, and the remaining 20\% (i.e., 292 samples) were used for testing. Similarly, there are also 80\% samples for training and 20\% for testing in the experiment of prescription recommendation.

In our proposed PrescDRL, the DQN network framework consists of three fully connected (FC) layers with 400, 300, and 30 neurons, respectively. For the DRQN network, the first two layers are FC layers with 300 and 512 neurons. The middle layer is an LSTM layer with 512 neurons, and the final layer is a FC layer with 30 neurons. Since there are 30 well-tuned HPCs, which correspond to 30 actions in modeling RL models, the DQN and DRQN layers have 30 neurons. During the training of these two models, the learning rate is 0.01, the discount coefficient of the reward value is 0.9, the random exploration probability is 0.1, and the batch size is 32. The parameters are copied to the Q-target network every 100 training batches.

\begin{figure}[!h]
\centering
\includegraphics[width=\linewidth]{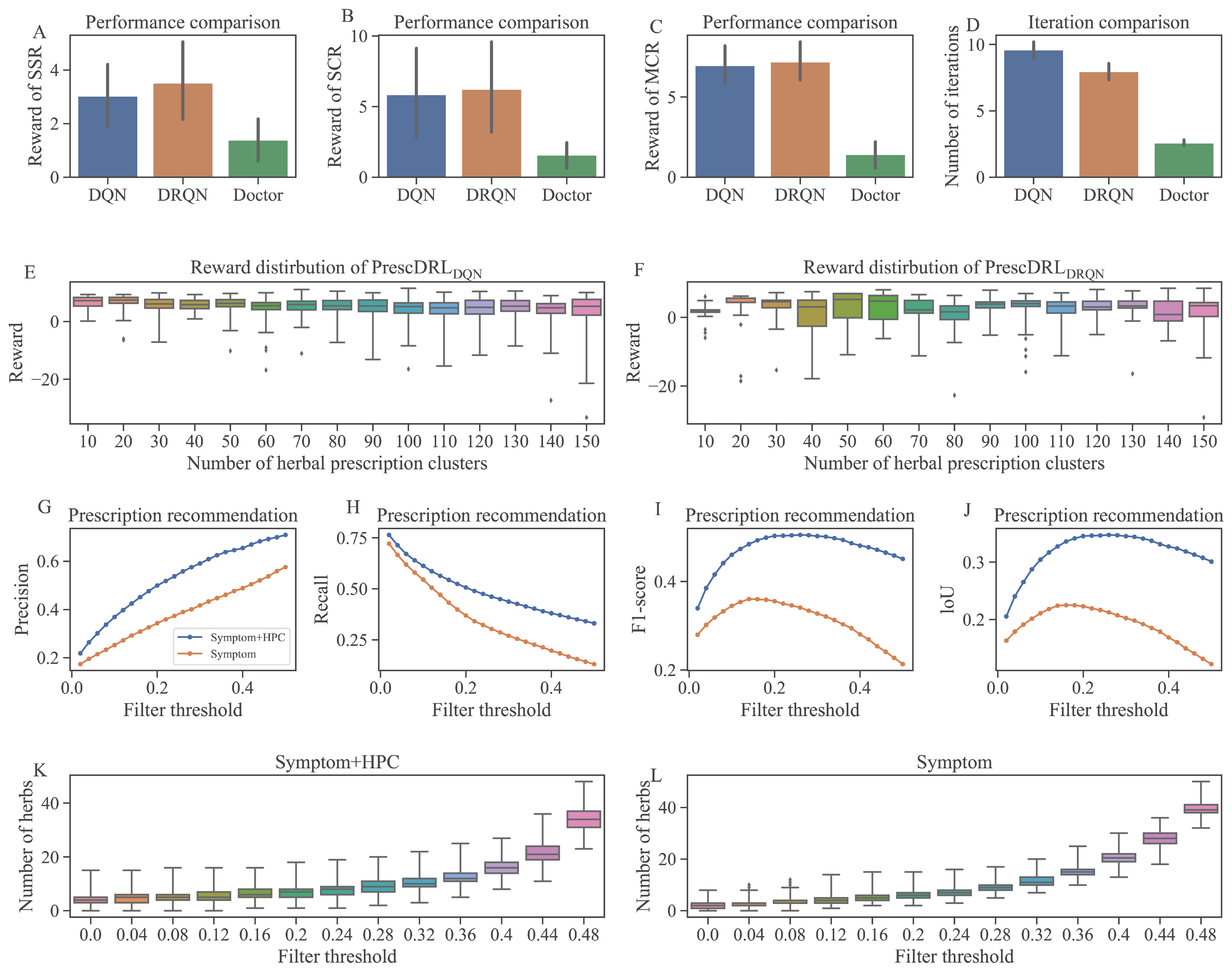}
\caption{Experimental results of PrescDRL. (A) Performance comparison of single-step return. (B) Performance comparison of single cumulative return. (C) Performance comparison of multi-step cumulative return. (D) Comparison of iterations, i.e., the sequence length of diagnosis and treatment plan. (E) Reward distribution of $PrescDRL_{DQN}$ with different HPCs. (F) Reward distribution of $PrescDRL_{DQN}$ with different HPCs. (G) Precision comparison of prescription recommendation. (H) Recall comparison of prescription recommendation. (I) F1-score comparison of prescription recommendation. (J) LoU comparison of prescription recommendation. (K) Number distribution of the predicted herbs given by PrescDRL that considers symptom and scheme. (L) Number distribution of the predicted herbs given by PrescDRL that only considers symptom.}
\label{fig:3}
\end{figure}

\subsubsection{Evaluation metrics}

In clinical practice, evaluating the effectiveness of TCM treatment for chronic diseases, such as diabetes, can be challenging due to the long duration of treatment and the unsuitability of western medicine's clinical mortality as an evaluation metric\cite{finkelstein1999combining}. In this study, we evaluated the performance of DDTS optimization results based on the improvement of symptom score, which is represented as the return values of RL models. To assess the effectiveness of the optimization models, we considered three commonly used metrics: single-step return (SSR), single-step cumulative return (SCR), and multi-step cumulative return (MCR). For SSR, the optimization models are trained based on the symptom observations of the first visit of each patient, and the differential between the symptom observations before and after the models provide an HPC is defined as SSR. In contrast, SCR considers all visits of each patient, and the average of all returns is computed. MCR is a more comprehensive metric, where the models predict an HPC based on the first visit of each patient and then use the state transition function to generate follow-up symptom observations until a set stopping condition is reached.

In addition, predicting prescriptions is considered a multi-label classification problem, for which precision, recall, F1 score, and IoU are used as evaluation metrics.

\section{RESULTS}

\subsection{Clustering and validation of diagnosis and treatment plan of diabetes}

Based on the PMET information of the prescriptions used by patients, we utilized the K-means clustering algorithm\cite{bellman1957markovian} to obtain prescription clusters. To determine the optimal number of HPCs, we conducted comprehensive experiments for parameter tuning and calculated the distribution of statistical difference (i.e., the negative logarithm of P-value) for each clustering result. A higher difference between categories implies better HPC results and indicates that the HPC obtained by clustering is more personalized. Analytical results showed that as the number of categories increases, the statistical difference of clustering results decreases (Figure~\ref{fig:2}C). We used a heatmap to illustrate the $-\log _{10} P$ distributions of the results of 10 and 150 categories, respectively. The statistical results indicated that the result of 30 categories had the highest statistical difference $\left(-\log _{10} P=23.27 \pm 8.27\right)$. We ultimately chose to cluster herbal prescriptions into 30 clusters as the action space of the deep RL model.

\subsection{Prescription treatment planning optimization of PrescDRL}
In this study, we proposed a deep RL-based method for predicting DDTS. To evaluate the performance of our proposed PrescDRL in predicting sequential HPCs, we compared the multiplex return values of the HPCs predicted by our model with those given by clinical doctors. In this experiment, we considered the results of 30 HPCs as the action space of PrescDRL. The comparison results (Figure~\ref{fig:3}A-\ref{fig:3}C and Table~\ref{table1}) reveal that PrescDRL obtains higher return values than clinical doctors on three evaluation metrics.

In terms of the single-step return (SSR), the clinical doctors achieved a score of 1.39, while $PrescDRL_{DQN}$ and $PrescDRL_{DRQN}$ improved by 117\% and 153\% compared to doctors, respectively. From the term of single-step cumulative return (SCR), compared to doctors, $PrescDRL_{DQN}$ and $PrescDRL_{DRQN}$ improved by 269\% and 292\% respectively. And for the multi-step cumulative return (MCR), $PrescDRL_{DQN}$ and $PrescDRL_{DRQN}$ improved by 387\% and 402\% than doctors respectively. Meanwhile, the results also showed $PrescDRL_{DRQN}$ obtain higher rewards than than $PrescDRL_{DQN}$, improved by 16.2\% for SSR, 6.48\% for SCR and 3.16\% for MCR.

We compared the length of diagnosis and treatment sequences of the PrescDRL model with that of doctors (Figure~\ref{fig:3}D). The results showed that doctors had the shortest sequence length, while both the $PrescDRL_{DQN}$ and $PrescDRL_{DRQN}$ models had longer sequence length than doctors. This indicates that although PrescDRL has high diagnostic performance, it increases the number of diagnosis and treatment. Additionally, we also compared the performance differences between $PrescDRL_{DQN}$ and $PrescDRL_{DRQN}$ models with different numbers of HPCs (Figure~\ref{fig:3}E and \ref{fig:3}F). The results showed that the number of HPCs has little influence on the prediction performance of PrescDRL. However, with the increase of HPC number, the $PrescDRL_{DRQN}$ model showed better stability than the $PrescDRL_{DQN}$ model.

\begin{table}[!h]
\footnotesize
\caption{Reward comparison of DDTS optimization.\label{table1}}%
\tabcolsep 49pt
\renewcommand{\arraystretch}{1.5}
\setlength{\tabcolsep}{4.5pt}
\begin{tabular*}{\textwidth}{@{\extracolsep{\fill}}lccc@{\extracolsep{\fill}}}
\toprule[1.2pt]
\rule{0pt}{15pt}
\textbf{Models} & \textbf{Single-step return} & \textbf{Single-step cumulative return} & \textbf{Multi-step cumulative return}\\
\midrule
$Doctor$ & 1.39 & 1.59 & 1.42 \\
$PrescDRL_{DQN}$ & 3.03 (117\% increase) & 5.86 (269\%) & 6.96 (387\%)\\
$PrescDRL_{DRQN}$ & 3.52 (153\%) & 6.24 (292\%) & 7.18 (402\%)\\
\bottomrule[1.2pt]
\end{tabular*}
\end{table}

The above results showed that our PrescDRL outperformed the doctors in terms of rewards, indicating better curative effects. Additionally, the sequential HPCs predicted by PrescDRL have shorter curative periods compared to those given by doctors. These results indicated that PrescDRL, based on RL, has significant advantages in DDTS optimization.

\subsection{Prescription prediction of PrescDRL}

In clinical practice, a prescription recommendation system needs to provide specific recommended prescriptions for each patient. However, PrescDRL provides an herbal prescription cluster rather than an actual herbal prescription. Therefore, we need to combine the HPC given by PrescDRL with the symptom observations of patients to make prescription predictions. The previous experimental results showed that the HPC provided by PrescDRL obtains higher rewards than those provided by clinicians. Therefore, the model that combines the HPC given by PrescDRL with symptom observations should recommend better prescriptions than doctors. Thus, it is not possible to evaluate the performance of PrescDRL using doctors’ prescriptions as a benchmark. To address this, we conducted a degradation experiment by combining the original HPC with symptom observations and used prescriptions given by doctors as a benchmark. If the prediction performance of this method is better than that using only symptom observation, it can be concluded that the prescription recommendation based on symptom observations and the HPC is better than that of doctors. Therefore, PrescDRL should have higher predictive performance.

\begin{table}[!h]
\footnotesize
\caption{Performance comparison of prescription recommendation\label{table2}}%
\tabcolsep 49pt
\renewcommand{\arraystretch}{1.5}
\setlength{\tabcolsep}{4.5pt}
\begin{tabular*}{\textwidth}{@{\extracolsep{\fill}}lcccccccc@{\extracolsep{\fill}}}
\toprule[1.2pt]
\rule{0pt}{15pt}
\textbf{Models} & \textbf{Filter threshold} & \textbf{No.of predicted herbs} & \textbf{Precision} & \textbf{Recall} & \textbf{F1-score} & \textbf{loU}\\
\midrule
\multirow{4}{*}{Symptom} & 0.1 & 3.77±1.34 & 0.49 & 0.20 & 0.28 & 0.17\\
\cline{2-7}
& 0.2 & 6.07±1.48 & 0.42 & 0.27 & 0.33 & 0.20\\
\cline{2-7}
& 0.3 & 10.12±1.77 & 0.34 & 0.37 & 0.36 & 0.22\\
\cline{2-7}
& 0.4 & 20.52±2.63 & 0.25 & 0.55 & 0.35 & 0.21\\
\midrule
\multirow{4}{*}{Symptom + scheme} & 0.1 & 5.44±2.39 & \textbf{0.65} & 0.38 & 0.48 & 0.33 \\
\cline{2-7}
& 0.2 & 7.04±2.60 & 0.59 & 0.44 & \textbf{0.50} & \textbf{0.34}\\
\cline{2-7}
& 0.3 & 9.72±2.80 & 0.50 & 0.51 & \textbf{0.50} & \textbf{0.34}\\
\cline{2-7}
& 0.4 & 16.00±3.03 & 0.37 & \textbf{0.61} & 0.46 & 0.30\\
\bottomrule[1.2pt]
\end{tabular*}
\end{table}

In the experiment, we constructed a three-layer fully connected neural network to compare prescription recommendation models. The first model considered both the patient's symptom information and the HPC given by PrescDRL, while the second model only considered the patient's symptom information and served as the benchmark. We evaluated the performance of these models under different filter thresholds and evaluation metrics (Figure~\ref{fig:3}G-\ref{fig:3}J and Table \ref{table2}). It's important to note that the filter threshold is a hyperparameter given to the neural network. The size of this parameter is inversely proportional to the strictness of the screening, meaning that smaller values correspond to stricter screening and larger values correspond to looser screening.

\begin{figure}[!b]
\centering
\includegraphics[width=\linewidth]{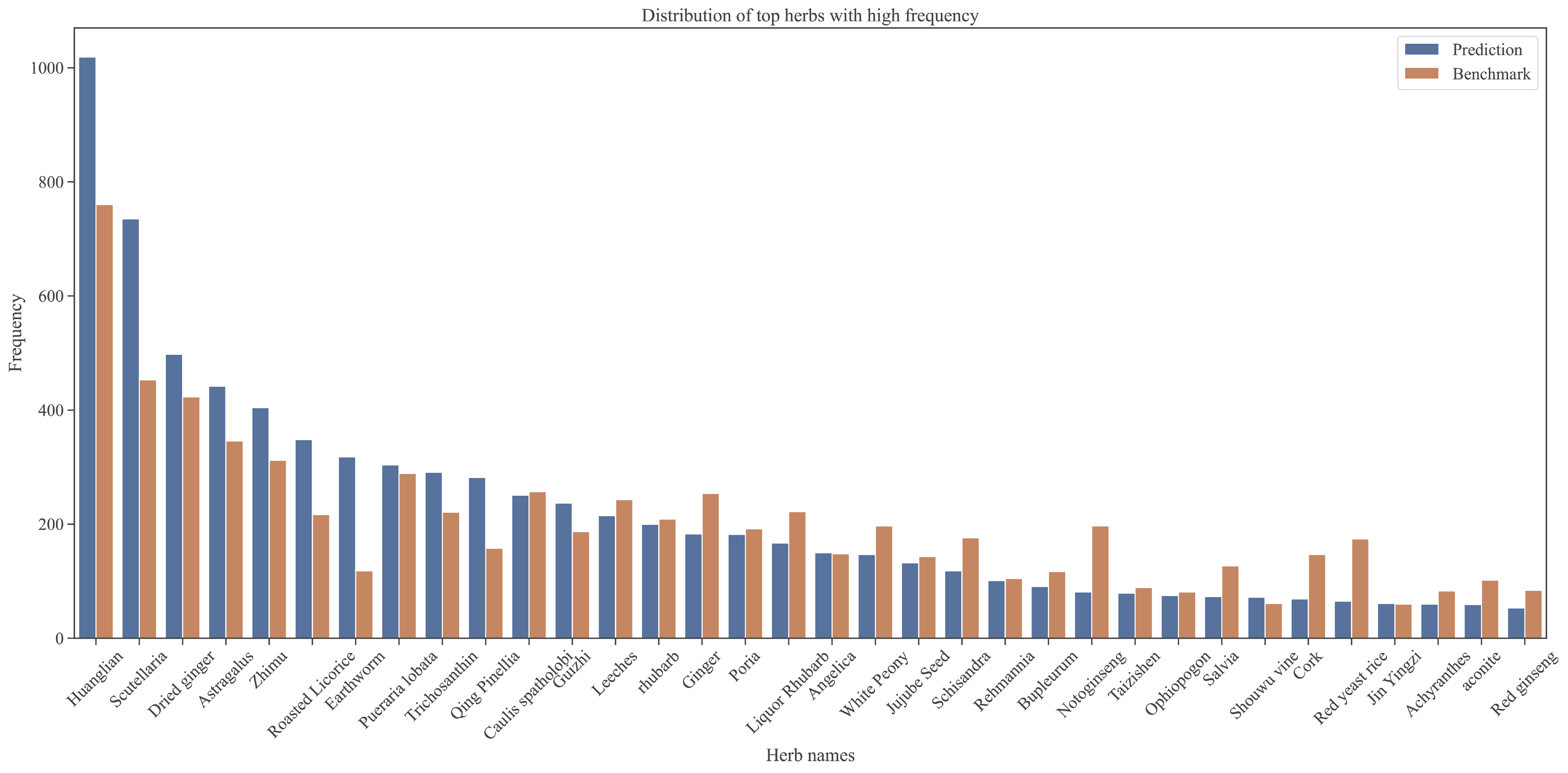}
\caption{Frequency distribution of top herbs predicted by PrescDRL.}
\label{fig:4}
\end{figure}

Our experimental results showed that the recommendation performance of our model was significantly higher than that of the benchmark model. For instance, with a filter threshold of 0.2, our model achieved higher precision (improved by 40.5\%), recall (improved by 63\%), and F1-score (improved by 51.5\%) than the benchmark model. Additionally, we compared the number of herbs recommended by the two models under different filter thresholds (Figure~\ref{fig:3}K-\ref{fig:3}L). The results showed that, as the filter threshold increased, the number of herbs recommended by both models increased. However, the number of herbs recommended by our model was lower than that of the benchmark model. When the filter threshold was set to 0.3, our model predicted an average of 9.72±2.80 herbs, while the benchmark model predicted an average of 10.12±1.77 herbs. Notably, in the real prescription data set, the mean and standard deviation of drugs in each prescription were 9.84±2.96. Thus, our model's predictions were closer to the true number of herbs in prescriptions when the appropriate filter threshold was selected.

\begin{table}[!b]
\footnotesize
\caption{Diagnosis and treatment sequence of doctors\label{table3}}%
\tabcolsep 49pt
\setlength{\tabcolsep}{0.1pt}
\begin{tabular*}{\textwidth}{@{\extracolsep{\fill}}ccccc@{\extracolsep{\fill}}}
\toprule[1.2pt]
\rule{0pt}{13pt}
\textbf{Sequence order} & \textbf{Symptom of patients} & \textbf{Herbal prescription cluster} & \textbf{Prescription} & \textbf{Symptom score} \\
\midrule
\multirow{4}{*}{1} & Tongue or coating with blood  & \multirow{4}{*}{67} & Coptis chinensis, donkey- & \multirow{4}{*}{13} \\
& stasis, dark tongue or coating, &  & hide gelatin beads, chicken \\
& rapid pulse, irregular stool, ...&  & yellow, white peony root, \\
& & &skullcap, jujube seed, ...\\
\rule{0pt}{16pt}
\multirow{4}{*}{2} & Thick and greasy tongue or & \multirow{4}{*}{37} & Dried ginger, Coptis  & \multirow{4}{*}{10} \\
& coating, stasis of tongue or  &  & chinensis, Scutellaria  \\
& coating, dark tongue or coating,&  & baicalensis, American \\ 
& aversion to cold, ...&  & ginseng, Poria, Panax \\ 
&  &  & notoginseng\\
\rule{0pt}{16pt}
\multirow{4}{*}{3} & The whole body is heavy,  & \multirow{4}{*}{67} & Angelica, Astragalus,  & \multirow{4}{*}{8} \\
& the tongue or fur is stagnated,  &  &  Coptidis, Cinnamon,   \\
& the tongue or fur is dark, &  & Anemarrhena, Golden \\ 
& the whole body is weak, ... &  & Cherry, Ginger, ... \\
\rule{0pt}{16pt}
\multirow{4}{*}{4} & Loose stools, stagnant tongue & \multirow{4}{*}{36} & wine rhubarb, aconite,   & \multirow{4}{*}{11} \\
& or coating, fluttering tongue &  & Alisma, fenugreek,    \\
&  or coating, thin tongue &  & Gorgon, yam, ...\\ 
& or coating, ...& & \\
\bottomrule[1.2pt]
\end{tabular*}
\end{table}

\begin{table}[h]
\footnotesize
\caption{Diagnosis and treatment sequence given by PrescDRL\label{table4}}%
\tabcolsep 49pt
\setlength{\tabcolsep}{4.5pt}
\begin{tabular*}{\textwidth}{@{\extracolsep{\fill}}ccccc@{\extracolsep{\fill}}}
\toprule[1.2pt]
\rule{0pt}{13pt}
\textbf{Inid} & \textbf{Symptom} & \textbf{Treatment plan} & \textbf{Prescription} & \textbf{Symptom score} \\
\midrule
\multirow{3}{*}{1} & Tongue or coating with blood  & \multirow{3}{*}{81} & Pueraria Root, & \multirow{3}{*}{13} \\
& stasis, dark tongue or coating, &  & Helichrysum,
Trichosanthes,\\
& rapid pulse, irregular stool, ...&  & Ginger,
Leeches, ...\\
\rule{0pt}{16pt}
\multirow{3}{*}{2} & Thick and greasy tongue or fur, & \multirow{3}{*}{81} & Astragalus, Salvia,
Zhigancao, & \multirow{3}{*}{8} \\
& stasis of tongue or fur, aversion  &  & Leech, Sophora Radix, \\
& to cold,dry or secretive stool, ...&  & Rhizoma Coptidis, ...\\ 
\rule{0pt}{16pt}
\multirow{4}{*}{3} & Swelling of lower extremities, & \multirow{4}{*}{12} & Treats, Astragalus, & \multirow{4}{*}{11} \\
& thick and greasy tongue or  &  &  Suanzaoren, Dilong,  \\
& coating, stasis of tongue &  & Sophora japonica, ...\\ 
& or coating, rapid pulse, ...& & \\
\rule{0pt}{16pt}
\multirow{4}{*}{4} & Tinnitus, dark tongue or & \multirow{4}{*}{71} & Pueraria, Monascus,  & \multirow{4}{*}{7} \\
& coating, sweating, yellow  &  & Dilong, Trichosanthes,   \\
& tongue or coating, &  &  Dried Ginger, Achyranthes, ...\\ 
& urination and nocturia, ...& & \\
\rule{0pt}{16pt}
5 & tinnitus, general fatigue & - & - & 2 \\

\bottomrule[1.2pt]
\end{tabular*}
\end{table}

\subsection{Case study of the diagnosis and treatment sequence of PrescDRL}
In the case study section, we first presented the distribution of herb occurrence frequency predicted by PrescDRL, with a filter threshold of 0.22, and compared the frequency of these herbs with those in the original prescriptions (Figure~\ref{fig:4}). The results showed that the predicted herbs had a similar frequency distribution to the original herbs, which indirectly confirmed the reliability of the predictive results generated by PrescDRL. 

To illustrate the predictive performance of PrescDRL, we showed a real diagnosis and treatment sequence (Table \ref{table3} and \ref{table4}) and a predicted sequence given by PrescDRL based on the first diagnosis of the real sequence. It is important to note that the two sequences are based on the results of 100 HPCs. Based on the doctor's sequence, the patient had a total of four visits with an initial symptom score of 13 and a symptom score of 11 for the last visit, indicating that the treatment effect was not optimal and there were recurrent conditions. This could be due to variations in patients' physical quality and medication, as well as differences in doctors' experience. Therefore, even if the same patient is at the same stage of disease, each doctor may prescribe different prescriptions according to their own experience.

After analyzing the diagnosis and treatment sequence provided by the PrescDRL algorithm, it was found that the patient had undergone a total of five visits and his symptom score had gradually decreased throughout the treatment, indicating a gradual improvement in the patient's condition. At the end of the sequence, the patient's symptom score was 2, which suggests that the patient's disease had significantly improved through the entire diagnosis and treatment process.

Through a comparison of the diagnosis and treatment sequences provided by the doctor and the PrescDRL algorithm, we observed that patient symptoms tend to improve in repeated fluctuations rather than directly. In the doctor's sequence, the symptom score gradually decreases in the first three visits, but then increases in the fourth visit, indicating that the fourth prescription may not have been appropriate. In contrast, the PrescDRL model aims to maximize long-term effectiveness by selecting the best medicine based on the patient's current symptoms, without necessarily expecting maximum reward at every step. As demonstrated in the sequence provided by PrescDRL, the symptom score does increase in the third visit, but ultimately drops to 2 in the fifth visit, suggesting that the diagnosis and treatment plan generated by the PrescDRL model based on reinforcement learning is more reasonable and effective.

\section{DISCUSSION}

With the advancement of real-world clinical medicine, a significant amount of diagnosis and treatment data from famous and experienced TCM doctors have been accumulated. As a result, how to extract medication rules from this data and develop an effective model for recommending reasonable prescriptions has become a research hotspot in TCM intelligence. In light of this, we propose a RL-based prediction model for optimizing diagnosis and treatment schemes. This model is a sequential optimization approach that prioritizes long-term effectiveness to provide rational TCM prescriptions. We designed the comprehensive experiments following the algorithm evaluation guidelines in the network pharmacology\cite{li2021network}, and the experimental results indicated that HPTP given by PrescDRL have better curative effect than doctors and higher performance on prescription prediction. 

There are still some areas that require further exploration in the future. First, the diagnosis and treatment sequences generated by PrescDRL are longer compared to those provided by doctors. In the future, it would be necessary to investigate a diagnosis and treatment optimization model that can provide shorter sequences while maintaining high efficacy. Second, all the experimental results presented in this study are based on simulation experiments. To validate the effectiveness of our proposed PrescDRL model, it is crucial to apply it in real-world clinical diagnosis and treatment systems to evaluate the specific effects.

\section{CONCLUSION}

In this study, we proposed PrescDRL, a deep RL-based optimization model for herbal prescription treatment planning, that prioritizes long-term effectiveness to provide reasonable TCM prescriptions. The experimental results demonstrated that PrescDRL generated herbal prescription treatment plans with better curative effects than doctors and achieved higher performance in herbal prescription recommendation. Overall, PrescDRL provides an exemplary approach to employ RL to learn the patient's optimal treatment path, which can help minimize medication errors, reduce the patient's treatment cost, and improve treatment effectiveness.





\section{CONTRIBUTIONS}

K.Y., X.Z. were involved in the conception and design of the work. Z.Y., X.H. and N.W. were involved in data collection and model construction. K.Y., Z.Y., X.H., Q.Z., F.Y. and X.S. were involved in data analysis and interpretation. K.Y., Z.L., T.W. and X.Z. were involved in the drafting and revision of the article. K.Y. and X.Z. approved the final version to be published.







\bibliographystyle{IEEEtran}
\bibliography{ref}






\end{document}